\documentclass{article}
\usepackage{arxiv}
\usepackage[T1]{fontenc}
\usepackage{graphicx}
\usepackage{subcaption}
\usepackage{booktabs}
\usepackage{wrapfig}
\usepackage[table]{xcolor}
\definecolor{rowgray}{gray}{0.85}
\definecolor{rowblue}{RGB}{218,232,252}
\usepackage{multirow}
\usepackage{adjustbox}
\usepackage[colorlinks,citecolor=blue]{hyperref}
\usepackage{orcidlink}
\usepackage{amsmath}
\usepackage{amssymb}
\usepackage{cleveref}
\usepackage{bm}

\begin{document}

\title{Stepwise Token Selection for Efficient Multimodal Large Language Models}

\author{
  Landi He \\
   Shenzhen University of Advanced Technology \\
  \And
  Shawn Young \\
 Shenzhen University of Advanced Technology \\
  \And
  Lijian Xu \thanks{Corresponding author.} \\
  Shenzhen University of Advanced Technology \\
}
\maketitle

\begin{abstract}
In multimodal large language models (MLLMs), inference cost is largely dominated by the visual token prefix rather than the language backbone, making token reduction a key factor for improving efficiency. Existing approaches typically assign independent importance scores to visual tokens and retain a fixed number of top-ranked tokens, implicitly assuming token independence and a uniform compression ratio across inputs.
In this work, we reformulate visual token pruning as a sequential decision-making process. Specifically, we introduce a pointer-style selection mechanism that iteratively chooses informative tokens, conditioning each decision on previously selected ones, and dynamically determines when to stop via a learned termination action. This enables joint optimization of both the selected subset and its size.
To enable end-to-end training under standard language modeling objectives, we design a differentiable relaxation based on a variance-preserving noise interpolation scheme, allowing gradients to propagate through the discrete selection process.
Extensive experiments on LLaVA-v1.5-7B and Qwen2.5-VL-7B demonstrate that our approach consistently outperforms fixed-ratio baselines across different compression levels. Under aggressive pruning that removes 88.9\% of visual tokens, our method preserves 94.6\% of the original accuracy while achieving a 1.88$\times$ speed-up in prefill latency.

\end{abstract}

\section{Introduction}
\label{sec:intro}

Multimodal large language models (MLLMs) such as LLaVA~\cite{llava}, Qwen-VL~\cite{qwenvl}, and GPT-4V~\cite{gpt4v} have become a standard architecture for vision--language reasoning. In these models the image is injected as a visual prefix: a frozen vision tower~\cite{clip} converts each image into hundreds of patch tokens, which are prepended to the text prompt and consumed by an autoregressive language model. Both inference latency and memory are dominated by the length of this prefix rather than by the language-model weights, because the per-layer cost of a transformer scales superlinearly with sequence length. Reducing the number of visual tokens that enter the language model is therefore one of the most effective ways to improve MLLM efficiency, and it has motivated a growing body of work on visual token compression~\cite{prunesurvey}.

Most existing approaches treat token pruning as a one-shot ranking problem: a relevance score is computed for every visual token, for example from in-LLM attention~\cite{fastv}, from \texttt{[CLS]} similarity~\cite{prumerge}, from vision-backbone attention dominance~\cite{visionzip}, or from cross-modal relevance to the text prompt~\cite{sparsevlm}, and a top-$K$ rule retains the $K$ highest-scoring tokens. Two assumptions underlie this approach. The first is independence: each token is scored on its own, so once a token is committed to the retained set, the marginal value of the remaining candidates, in particular duplicates or close neighbours of the committed one, is not re-estimated even though it has changed. The second is a fixed compression ratio: $K$ is chosen externally and applied uniformly, regardless of whether the input is a near-empty document thumbnail or a dense natural scene. Real inputs vary widely in visual complexity, and no single $K$ is appropriate for all of them. We argue that both assumptions should be dropped, and that token selection is better framed as a stepwise sequential decision process in which the model commits to one token per step, conditions each step on tokens chosen earlier, and decides for itself when to stop.

\begin{figure}[t]
\centering
\includegraphics[width=\textwidth]{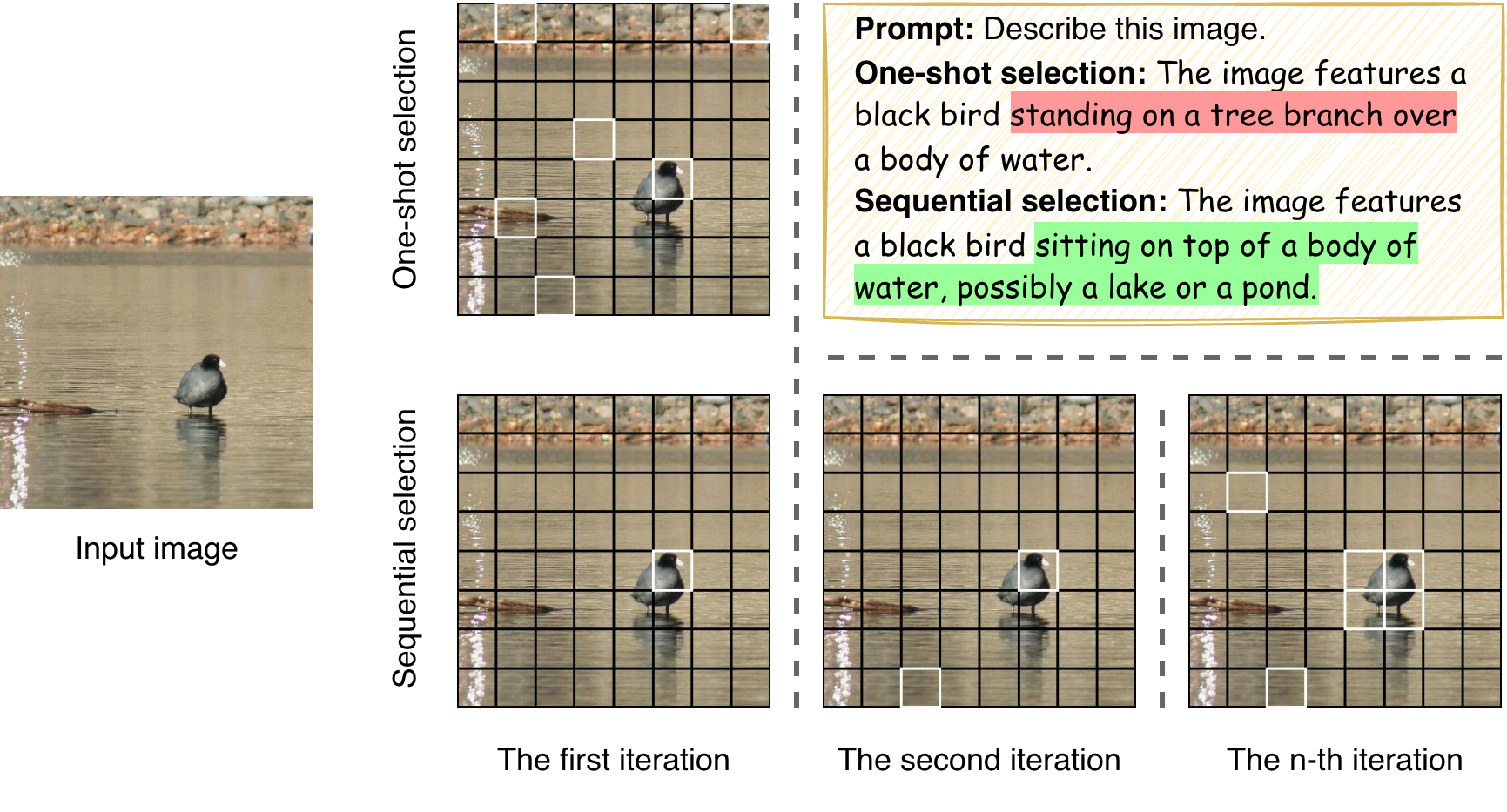}
\caption{One-shot scoring (top) versus stepwise sequential selection (bottom). Each row shows the retained tokens and the caption that the LLM produces from the resulting visual prefix.}
\label{fig:motivation}
\end{figure}

Motivated by this view, we propose StepPrune, which casts selection as an autoregressive policy $P(\mathcal{S}\mid\mathbf{X}) = \prod_t \pi_\theta(a_t\mid a_{<t},\mathbf{X})$ realised by a pointer-style decoder. The candidate set augments the visual tokens with a learned stop action $\varnothing$, and the episode terminates the first time $\varnothing$ is chosen, so the retained count $K$ becomes an adaptive, per-input quantity rather than a fixed hyperparameter. Because the per-step argmax is non-differentiable, we pass the aggregated pointer probabilities through a variance-preserving noise gate that mixes each visual token with isotropic Gaussian noise under a smooth polarisation of its selection score, while a locally-masked denoiser prevents cross-token leakage from bypassing the gating signal. Figure~\ref{fig:motivation} contrasts the two regimes, where a one-shot prefix drops spatial cues that the language model then hallucinates around, whereas the stepwise policy preserves the evidence needed for a correct description.

In summary, the contributions of this paper are threefold.
\begin{itemize}
\item We formulate visual-token pruning for MLLMs as a conditional sequential decision process with a learned stop action, which to our knowledge has not been done in this form before, in place of the one-shot scoring-plus-top-$K$ paradigm.
\item The retained count $K$ becomes an adaptive, per-input quantity produced by the policy under a light length prior, so no target compression ratio needs to be specified at inference time.
\item A deterministic variance-preserving noise gate, combined with score polarisation and a diagonally-masked denoiser, makes the sequential argmax trainable end-to-end from the language-modelling loss. On standard MLLM benchmarks, StepPrune preserves $97.9\%/96.7\%/94.6\%$ of the full-model accuracy at $66.7\%/77.8\%/88.9\%$ pruning rates and yields a $1.88\times$ prefill speed-up at the most aggressive setting, outperforming fixed-ratio baselines at every operating point (Sec.~\ref{sec:exp}).
\end{itemize}

\section{Related Work}
\label{sec:related}

\subsection{Multimodal Large Language Models}
\label{sec:related:mllm}
Modern MLLMs~\cite{llava,blip2,young2026scalar,xu2026unified,xu2024foundation,young2026xrayclaw} couple a frozen vision tower with a generative language model through a learned projection. As resolutions grow, the visual prefix routinely occupies several hundred patch tokens, and recent surveys~\cite{prunesurvey} identify its length, rather than the language-model weights, as the main remaining target for inference efficiency. 
In medical imaging domains, multimodal foundation models have been extensively explored for pathology and radiology analysis~\cite{yang2024segmentation,yang2023geometry,feng2026efficient,wu2026multimodal,xu2024medvilam,xu2026unified,xu2024foundation,chen2026tc}, which demonstrate strong representation learning capabilities but still suffer from the same long visual token bottleneck in high-resolution inputs.
We organise the remainder by how prior methods produce the prefix-shrinking decision.

\subsection{Heuristic Visual Token Pruning}
\label{sec:related:heuristic}
A first family of methods computes a closed-form relevance score for every visual token and keeps the top-$K$ ranked entries, with $K$ supplied externally as a fixed hyperparameter. LLaVA-PruMerge~\cite{prumerge} uses \texttt{[CLS]}-similarity, FastV~\cite{fastv} uses in-LLM attention statistics, and SparseVLM~\cite{sparsevlm} uses cross-modal relevance to the text prompt. A complementary line of work prioritises representational coverage over raw saliency: VisionZip~\cite{visionzip} keeps tokens that dominate the vision tower's own attention, and DART~\cite{dart} selects explicitly against duplication. Across this design space, every method shares the two assumptions identified in Sec.~\ref{sec:intro}: each token is scored independently of the others, and the budget $K$ is held constant across inputs.

\subsection{Learned and End-to-End Token Selection}
\label{sec:related:learned}
A more recent line of work makes the pruning decision a learned function rather than a closed-form score, so that pruning behaviour can depend on each input ~\cite{he2026autoselect,gao2026zerosense}. GlimpsePrune~\cite{glimpseprune} trains a token-importance predictor under bounding-box supervision, which ties the selection signal to a localisation surrogate. ATP-LLaVA~\cite{atpllava} interleaves adaptive prediction modules between language-model layers to estimate per-instance retention ratios, relaxing hard selection through differentiable approximations but relying on auxiliary losses to keep the implicit budget in check. p-MoD~\cite{pmod} recasts visual-token retention as a Mixture-of-Depths routing problem and obtains an adaptive depth per token, at the cost of jointly fine-tuning the entire language model. Concurrent work such as TopV~\cite{topv} pursues similar goals under different optimisation surrogates. These methods relax the fixed-$K$ assumption, but each retained token is still produced by a per-token decision that does not depend on which other tokens have already been chosen, so the independence assumption inherited from one-shot ranking remains.

StepPrune differs from this body of work along the conditioning axis. We model the retained set $\mathcal{S}$ as the trajectory of an autoregressive policy whose candidate space includes a learned stop action $\varnothing$, so that which tokens are retained and how many are retained are decided jointly. Two ideas from prior work are combined to make this concrete. The decoder reuses the pointer-attention mechanism of Pointer Networks~\cite{pointernet}, applied here not to a combinatorial task with a fixed-length output but to a variable-length subset and a stop event drawn from the same pointer distribution. The differentiable training pathway adapts the variance-preserving noise gate of~\cite{autoprune} to handle the discrete argmax; in our setting, the gate of token $i$ is the marginal of the sequential policy, and is therefore a function of every preceding pointer step rather than a single-shot quantity. We are not aware of a prior MLLM token-pruning method that folds both the membership of $\mathcal{S}$ and its cardinality $K$ into the same conditional sequential decision.

\section{Method}
\label{sec:method}

We formulate visual token pruning for multimodal large language models (MLLMs) as a stepwise sequential decision problem. Instead of scoring all visual tokens in a single forward pass, StepPrune selects one token per step, with each step conditioned on the tokens chosen earlier. Sec.~\ref{sec:method:prelim} fixes the notation, Sec.~\ref{sec:method:decoder} describes the conditional pointer decoder, Sec.~\ref{sec:method:training} introduces the noise-gating pathway used for end-to-end training, and Sec.~\ref{sec:method:loss} states the training objective.

\subsection{Preliminaries}
\label{sec:method:prelim}

\paragraph{Setup and Task.}
A LLaVA-style MLLM encodes an input image with a vision tower, e.g., CLIP ViT-L/14, into $\mathbf{X}=[\mathbf{x}_1,\dots,\mathbf{x}_N]\in\mathbb{R}^{N\times D}$ with $N{=}576$ and $D{=}1024$. The result is projected and prepended to the text prompt embeddings $\mathbf{E}^{\text{txt}}\in\mathbb{R}^{L_{\text{txt}}\times D_{\text{txt}}}$ before being passed to an autoregressive LLM. Token pruning seeks an ordered subset $\mathcal{S}\subseteq\{1,\dots,N\}$ of size $K{=}|\mathcal{S}|$ that minimises the downstream loss when the LLM consumes $\mathbf{X}_{\mathcal{S}}$ in place of $\mathbf{X}$. Existing methods commonly express this as an independent scoring problem,
\begin{equation}
\label{eq:indep_score}
s_i = f_{\phi}(\mathbf{x}_i,\,\mathrm{ctx}),\qquad \mathcal{S} = \mathrm{TopK}(\{s_i\}_{i=1}^{N},\,K),
\end{equation}
which cannot capture the fact that, once token $i$ is committed to $\mathcal{S}$, the marginal value of every remaining candidate is changed by the information that $i$ already contributes. We therefore reformulate selection as a sequential decision process,
\begin{equation}
\label{eq:seq_factorization}
P(\mathcal{S}\mid\mathbf{X}) \;=\; \prod_{t=1}^{T}\,\pi_{\theta}\!\left(a_t\,\big|\,a_{<t},\mathbf{X}\right),
\end{equation}
where $a_t\in\{1,\dots,N\}\cup\{\varnothing\}$ is the action at step $t$ and $\varnothing$ denotes a learned \emph{stop} action. The episode terminates at the first $t$ with $a_t=\varnothing$, denoted $T^{*}$, and $K{=}T^{*}{-}1$ is determined by the model rather than fixed as a hyperparameter. The pointer distribution at step $t$ over the $N{+}1$ candidates is denoted $\mathbf{p}_t$.

\subsection{Conditional Pointer Decoder}
\label{sec:method:decoder}

\begin{figure}[t]
\centering
\includegraphics[width=\textwidth]{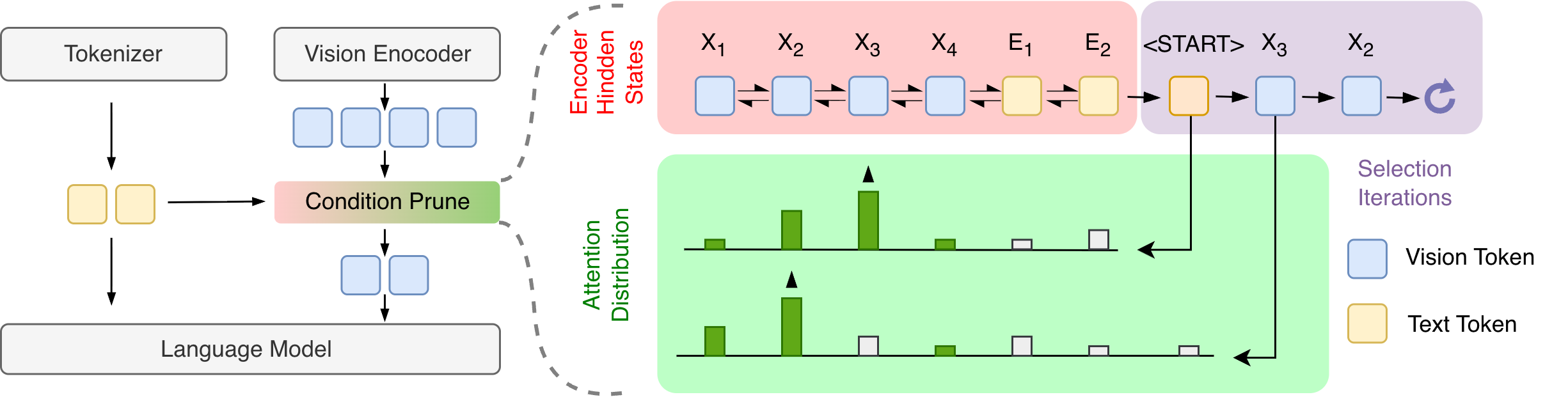}
\caption{Architecture of StepPrune. (Left) Module placement in a LLaVA-style pipeline. (Right) Internal mechanism: image patch tokens (blue) and projected text embeddings (yellow) are jointly encoded into the decoder memory $\mathbf{M}$ under bidirectional self-attention, after which a causally-masked pointer decoder queries $\mathbf{M}$ across selection iterations to emit pointer distributions over the $N{+}1$ candidates. The two attention patterns reflect the asymmetry of the task: encoding sees the full image at once, while selection is autoregressive over its own history.}
\label{fig:arch}
\end{figure}

StepPrune implements the policy $\pi_{\theta}$ in Eq.~\eqref{eq:seq_factorization} using three stacked modules: a cross-modal input encoder, a pointer decoder, and a learned stop token that augments the candidate set. Fig.~\ref{fig:arch} shows the overall architecture, and the modules are described in turn below.

\paragraph{Cross-Modal Input Encoder.}
The encoder consists of two pre-norm Vision Transformer blocks. To make the selection policy aware of the user instruction, the LLM word embeddings $\mathbf{E}^{\text{txt}}$ are projected into the visual space and concatenated with the image tokens before joint self-attention,
\begin{equation}
\label{eq:text_inject}
\tilde{\mathbf{E}}^{\text{txt}} \;=\; g_{\text{txt}}\cdot \mathrm{LN}\!\left(\mathbf{W}_{p}\,\mathbf{E}^{\text{txt}}\right),
\qquad
\mathbf{H} \;=\; \mathrm{Enc}\!\left([\,\mathbf{X};\,\tilde{\mathbf{E}}^{\text{txt}}\,]\right)_{[1{:}N]}\in\mathbb{R}^{N\times D},
\end{equation}
where $\mathbf{W}_{p}\in\mathbb{R}^{D\times D_{\text{txt}}}$ is a learned projection and $g_{\text{txt}}\in[0,1]$ is a scheduled gate that anneals from $0$ to $1$ during early training to stabilise optimisation. After encoding, only the image positions are retained as decoder memory: the text tokens have already shaped $\mathbf{H}$ through cross-modal attention, and selection is restricted to image tokens.

\paragraph{Memory and the Stop Action.}
A learned vector $\mathbf{s}_{\text{stop}}\in\mathbb{R}^{D}$ is appended to $\mathbf{H}$ to form the decoder memory
\begin{equation}
\label{eq:memory}
\mathbf{M} \;=\; [\,\mathbf{H};\,\mathbf{s}_{\text{stop}}\,]\in\mathbb{R}^{(N+1)\times D}.
\end{equation}
Treating the stop action as an additional candidate makes ``when to halt'' a learned property of the selection policy, so no target compression ratio needs to be specified at inference time.

\paragraph{Pointer Decoder Layer.}
We adopt the pointer attention mechanism of Vinyals et al.~\cite{pointernet} as the selection primitive. Each decoder layer consists of (i) causal self-attention over previously chosen embeddings, which propagates conditioning across selection steps, (ii) cross-attention from the current query to the memory $\mathbf{M}$, whose attention logits are reused as pointer logits, and (iii) a feed-forward network.

\paragraph{Autoregressive Selection Loop.}
The decoder is initialised with a learned start token $\mathbf{q}_1{=}\mathbf{s}_{\text{start}}$ and an empty log-space mask $\mathrm{Mask}_1\equiv\mathbf{0}\in\mathbb{R}^{N+1}$; the loop proceeds for at most $T_{\max}$ steps. At step $t$,
\begin{align}
\label{eq:ptr_logits}
\boldsymbol{\ell}_t &\;=\; \mathrm{PtrAttn}\!\left(\mathbf{q}_t,\mathbf{M}\right)\;+\;\mathrm{Mask}_t,\\
\label{eq:ptr_prob}
\mathbf{p}_t &\;=\; \mathrm{softmax}\!\left(\boldsymbol{\ell}_t\,/\,\tau\right),\\
\label{eq:ptr_argmax}
a_t &\;=\; \arg\max_{j\in\{1,\dots,N+1\}} \boldsymbol{\ell}_t(j),\\
\label{eq:ptr_mask}
\mathrm{Mask}_{t+1}(i) &\;=\; \mathrm{Mask}_t(i) \;+\; (-\infty)\cdot\mathbb{1}[\,a_t = i\,],
\end{align}
where $\tau$ is a softmax temperature. The selected memory slot is fed back as the next decoder query through a straight-through estimator, $\mathbf{q}_{t+1}{=}\mathbf{M}^{\!\top}\!\big(\,\mathrm{onehot}(a_t)\,-\,\mathrm{sg}(\mathbf{p}_t)\,+\,\mathbf{p}_t\,\big)$, where $\mathrm{sg}(\cdot)$ is the stop-gradient operator, so the forward pass uses the hard selection while gradients still flow through $\mathbf{p}_t$. The episode terminates when $a_t{=}\varnothing$ or $t{=}T_{\max}$, yielding a per-sample selection set $\mathcal{S}{=}\{a_1,\dots,a_{T^{*}-1}\}$ of adaptive size $K$. The joint probability decomposes as
\begin{equation}
\label{eq:joint}
P(a_{1{:}T}\mid\mathbf{X}) \;=\; \prod_{t=1}^{T}\mathbf{p}_t(a_t),
\end{equation}
which is the form of Eq.~\eqref{eq:seq_factorization} used in practice.

\subsection{Differentiable Selection via Noise-Gated Tokens}
\label{sec:method:training}

The argmax in Eq.~\eqref{eq:ptr_argmax} is non-differentiable, and a REINFORCE estimator on Eq.~\eqref{eq:joint} has high variance when $T$ is on the order of hundreds. We therefore train the policy through a deterministic surrogate that rewrites the discrete selection as a continuous, variance-preserving perturbation of the original tokens.

\paragraph{Aggregated Soft Scores.}
Let $A_t$ denote the probability that the episode has not terminated before step $t$,
\begin{equation}
\label{eq:alive_prob}
A_t \;=\; \prod_{u=1}^{t-1}\bigl(1-p_u(\varnothing)\bigr),\qquad A_1 = 1.
\end{equation}
For each visual token $i$, we accumulate the geometrically-discounted, alive-weighted pointer probabilities over all decoding steps:
\begin{equation}
\label{eq:soft_score}
s_i \;=\; \sum_{t=1}^{T_{\max}} A_t\cdot p_t(i)\cdot \beta^{t},
\qquad \beta\in(0,1).
\end{equation}
The score $s_i$ is high when token $i$ is pointed at with large probability early in the trajectory and before the policy has likely terminated; tokens that are never pointed at receive scores close to zero. Because $s_i$ depends smoothly on every $\mathbf{p}_t$ and therefore on $\theta$, any downstream loss expressed through $s_i$ supports standard back-propagation.

\paragraph{Polarisation.}
The raw scores in Eq.~\eqref{eq:soft_score} aggregate pointer mass across selection steps, but their absolute scale is governed by $T_{\max}$ and $\beta$ rather than by the discrete selection they are meant to approximate. A second softening is therefore applied to remap them onto a bounded range that resembles a hard top-$k$ indicator. We feed $\mathbf{s}$ through a differentiable top-$k$ polarisation operator~\cite{softtopk},
\begin{equation}
\label{eq:polarize}
\alpha_i \;=\; \mathrm{SoftTopK}_{k,\tau}\!\left(\mathbf{s}\right)_{i}\,\in\,[0,1],
\qquad k = \mathbb{E}_{b}\!\left[K_{b}\right],
\end{equation}
where $k$ is set to the running batch-average of the hard count $K$ and $\tau$ controls the sharpness. SoftTopK is used as a smooth re-mapping of scores into $[0,1]$; the actual selection is performed by the discrete pointer in Eq.~\eqref{eq:ptr_argmax}.

\paragraph{Variance-Preserving Noise Gating.}
With $\alpha_i$ in hand, we obtain a continuous proxy for the discrete selection by mixing each token with isotropic Gaussian noise:
\begin{equation}
\label{eq:vp_noise}
\tilde{\mathbf{x}}_i \;=\; \sqrt{\alpha_i}\,\mathbf{x}_i \;+\; \sqrt{1-\alpha_i}\,\boldsymbol{\epsilon}_i,
\qquad \boldsymbol{\epsilon}_i \sim \mathcal{N}(\mathbf{0},\mathbf{I}_D).
\end{equation}
This is the variance-preserving noise gate of AutoSelect~\cite{autoprune}, used here to connect the selection probabilities with the language-modelling loss: $\alpha_i\!\to\!1$ recovers the original token, $\alpha_i\!\to\!0$ replaces it with pure noise, and the second moment of $\tilde{\mathbf{x}}_i$ is preserved across the entire $[0,1]$ range. The main difference from~\cite{autoprune} is the source of $\alpha_i$: rather than a per-token independent score, our gate is the marginal of a sequential decision policy, so the gate of token~$i$ depends on every other token's preceding probability. The autoregressive per-step selection together with the learned stop action gives StepPrune its name: the retained set is built up one step at a time, and the policy decides when no further step is warranted.

\paragraph{Local Denoiser.}
Direct application of Eq.~\eqref{eq:vp_noise} injects high-variance noise into the spatial token grid. To counteract the resulting drift while preventing the denoiser from learning a shortcut that bypasses the gating signal, we apply a single Vision Transformer block whose self-attention is restricted to the diagonal,
\begin{equation}
\label{eq:denoise}
\hat{\mathbf{X}} \;=\; \mathrm{Denoise}\!\left(\tilde{\mathbf{X}};\,\mathrm{mask}=\mathrm{diag}\right).
\end{equation}
The diagonal mask permits a per-token MLP-style update through the block's residual path but blocks all cross-token information flow, so a noised token cannot recover its content from its neighbours.

\paragraph{Train--Inference Alignment.}
At training time, the language model consumes the full-length noise-gated $\hat{\mathbf{X}}\in\mathbb{R}^{N\times D}$ in its original spatial order; at inference time it consumes the hard subset $\mathbf{X}_{\mathcal{S}}\in\mathbb{R}^{K\times D}$ produced by Eqs.~\eqref{eq:ptr_logits}--\eqref{eq:ptr_mask}. The two pathways agree in expectation: as the polarisation in Eq.~\eqref{eq:polarize} sharpens, $\alpha_i\!\to\!\mathbb{1}[i\in\mathcal{S}]$ and Eq.~\eqref{eq:vp_noise} reduces to a hard pass-through for selected tokens.

\subsection{Training Objective}
\label{sec:method:loss}

The denoised noise-gated token sequence $\hat{\mathbf{X}}$ is passed through the standard LLaVA visual projector and concatenated with the prompt embeddings to compute the autoregressive language-modelling loss $\mathcal{L}_{\text{LM}}(\hat{\mathbf{X}})$. To bias the policy towards compression without prescribing a target ratio, we add a length penalty equal to the realised selection ratio,
\begin{equation}
\label{eq:loss}
\mathcal{L} \;=\; \mathcal{L}_{\text{LM}}\!\left(\hat{\mathbf{X}}\right) \;+\; \lambda\cdot\frac{K}{N},
\qquad \lambda = 10^{-2}.
\end{equation}
Because $K$ is determined by the first occurrence of the stop action $\varnothing$, the gradient of the length term flows through $\{p_t(\varnothing)\}_{t=1}^{T_{\max}}$ in Eq.~\eqref{eq:alive_prob} back to both the stop-token embedding $\mathbf{s}_{\text{stop}}$ and every decoder parameter, providing a soft learning pressure that competes with the language-modelling loss. The vision tower and the language model are kept frozen throughout; only the parameters of $\mathrm{Enc}$, $\mathrm{Dec}_{\theta}$, $\mathrm{Denoise}$, $\mathbf{W}_{p}$, $\mathbf{s}_{\text{start}}$, and $\mathbf{s}_{\text{stop}}$ are updated. Optimiser settings, the schedule for $g_{\text{txt}}$, and the value of $T_{\max}$ are given in Sec.~\ref{sec:exp}.

\section{Experiments}
\label{sec:exp}

We evaluate StepPrune on standard multimodal benchmarks against state-of-the-art visual-token pruning methods, using two backbones of different scales to test whether the conditional policy transfers beyond a single architecture. In all experiments, the vision tower and the language model are kept frozen, and only the encoder, decoder, denoiser, pointer projection, and the two learnable special tokens introduced in Sec.~\ref{sec:method} are trained.

\subsection{Experimental Setup}
\label{sec:exp:setup}

\paragraph{Benchmarks.}
We evaluate on eight standard multimodal benchmarks: GQA~\cite{gqa}, MMBench and MMBench-CN~\cite{mmbench}, MME~\cite{mme}, POPE~\cite{pope}, ScienceQA-IMG~\cite{scienceqa}, VQA$^{\text{v2}}$~\cite{vqav2}, and TextVQA~\cite{textvqa}, which together cover general VQA, multilingual reasoning, science QA, text-rich understanding, and hallucination robustness. The protocol follows FastV~\cite{fastv} and SparseVLM~\cite{sparsevlm} for direct comparability with prior work.

\paragraph{Baselines.}
We compare against four representative families of visual-token pruning methods. FastV~\cite{fastv} represents attention-based pruning and ranks tokens by accumulated attention inside the language model. LLaVA-PruMerge~\cite{prumerge} represents CLS-similarity pruning and scores every patch token against the vision backbone's \texttt{[CLS]} summary. VisionZip~\cite{visionzip} represents vision-dominance pruning and selects tokens that dominate the vision tower's own attention. SparseVLM~\cite{sparsevlm} represents text-conditioned pruning and re-weights tokens by cross-modal relevance to the prompt. Together, these four baselines cover the main mechanisms catalogued in the recent survey of multimodal token compression~\cite{prunesurvey}.

\paragraph{Implementation details.}
Unless stated otherwise, our primary backbone is LLaVA-v1.5-7B~\cite{llava} with the CLIP-ViT-L/14~\cite{clip} vision tower, which produces $N{=}576$ visual tokens per image. We also report transfer to Qwen2.5-VL-7B~\cite{qwenvl} in Sec.~\ref{sec:exp:qwen}. The trainable modules are the input encoder $\mathrm{Enc}$, the pointer decoder $\mathrm{Dec}_{\theta}$, the diagonally-masked denoiser $\mathrm{Denoise}$, the pointer projection $\mathbf{W}_{p}$, and the two learnable tokens $\mathbf{s}_{\text{start}}$ and $\mathbf{s}_{\text{stop}}$ defined in Sec.~\ref{sec:method}; the vision tower and the LLM are frozen. We set $T_{\max}{=}\lfloor N/2\rfloor{=}288$, the geometric decay base $\beta{=}0.9$, and the length-penalty weight $\lambda{=}10^{-2}$. The text-gating coefficient $g_{\text{txt}}$ is annealed from $0$ to $1$ over the course of training. We use AdamW with a cosine learning-rate schedule decaying from $5{\times}10^{-6}$ to $5{\times}10^{-7}$, and an effective global batch size of $16$ after gradient accumulation, on $2{\times}$\,NVIDIA RTX A6000 GPUs.

\subsection{Main Results on LLaVA-v1.5-7B}
\label{sec:exp:main_llava}

Table~\ref{tab:main_llava} reports StepPrune against the four baselines on LLaVA-v1.5-7B at three retention budgets, $K{=}192/128/64$, which correspond to $66.7\%/77.8\%/88.9\%$ pruning. Block labels denote a per-image fixed budget for the baselines; for StepPrune, $K$ varies per input, so the label reports the test-split mean $\bar{K}$ produced by the adaptive policy.

\begin{table}[t]
\centering
\caption{
  Results on LLaVA-v1.5-7B at three retention budgets, with the full-prefix vanilla model as upper bound. ``Avg.'' is the mean ratio of each benchmark score to the upper bound. For Avg., the best is in \textcolor{red}{red} and the second best in \textcolor{blue}{blue}; 
  Our method is highlighted in color.
}
\label{tab:main_llava}
\scriptsize
\setlength{\tabcolsep}{3pt}
\resizebox{\linewidth}{!}{%
\begin{tabular}{lccccccccc}
\toprule
Method & GQA & MMB & MMB$_{\text{CN}}$ & MME & POPE & SQA & VQA$^{\text{v2}}$ & VQA$^{\text{Text}}$ & Avg. \\
\midrule
\rowcolor{rowgray} \multicolumn{10}{c}{\textit{Upper Bound, 576 Tokens (100\%)}} \\
Vanilla & 61.9 & 64.7 & 58.1 & 1862 & 85.9 & 69.5 & 78.5 & 58.2 & 100\% \\
\midrule
\rowcolor{rowgray} \multicolumn{10}{c}{\textit{Retain 192 Tokens (66.7\% pruning)}} \\
FastV~{\scriptsize\textcolor{gray}{(ECCV'24)}}     & 52.7 & 61.2 & 53.5 & 1612 & 64.8 & 67.3 & 67.1 & 52.5 & 88.3\% \\
SparseVLM~{\scriptsize\textcolor{gray}{(ICML'25)}} & 57.6 & 62.5 & 58.6 & 1721 & 83.6 & 69.1 & 75.6 & 56.1 & 96.5\% \\
PDrop~\cite{pdrop}{\scriptsize\textcolor{gray}{(CVPR'25)}}     & 57.3 & 63.6 & 56.8 & 1797 & 82.3 & 69.2 & 75.1 & 56.5 & 96.0\% \\
VisionZip~{\scriptsize\textcolor{gray}{(CVPR'25)}} & 59.3 & 63.0 & 57.3 & 1783 & 85.3 & 68.9 & 76.8 & 57.3 & \textcolor{blue}{97.8}\% \\
\rowcolor{rowblue!40} StepPrune (Ours) & 59.5 & 63.5 & 57.4 & 1773 & 85.7 & 69.9 & 76.6 & 56.5 & \textcolor{red}{97.9\%} \\
\midrule
\rowcolor{rowgray} \multicolumn{10}{c}{\textit{Retain 128 Tokens (77.8\% pruning)}} \\
FastV~{\scriptsize\textcolor{gray}{(ECCV'24)}}     & 49.6 & 56.1 & 55.9 & 1490 & 59.6 & 60.2 & 61.8 & 50.6 & 85.2\% \\
SparseVLM~{\scriptsize\textcolor{gray}{(ICML'25)}} & 56.0 & 60.0 & 51.1 & 1696 & 80.5 & 67.1 & 73.8 & 54.9 & 92.4\% \\
PDrop~{\scriptsize\textcolor{gray}{(CVPR'25)}}     & 57.1 & 61.6 & 56.6 & 1761 & 82.3 & 68.4 & 72.9 & 56.6 & 94.7\% \\
VisionZip~{\scriptsize\textcolor{gray}{(CVPR'25)}} & 57.6 & 62.0 & 56.7 & 1762 & 83.2 & 68.9 & 75.6 & 56.8 & \textcolor{blue}{96.4}\% \\
\rowcolor{rowblue!40} StepPrune (Ours) & 57.3 & 63.2 & 57.1 & 1749 & 85.2 & 69.8 & 75.7 & 55.3 & \textcolor{red}{96.7\%} \\
\midrule
\rowcolor{rowgray} \multicolumn{10}{c}{\textit{Retain 64 Tokens (88.9\% pruning)}} \\
FastV~{\scriptsize\textcolor{gray}{(ECCV'24)}}     & 46.1 & 48.0 & 52.7 & 1256 & 48.0 & 51.1 & 55.0 & 47.8 & 76.8\% \\
SparseVLM~{\scriptsize\textcolor{gray}{(ICML'25)}} & 52.7 & 56.2 & 46.1 & 1505 & 75.1 & 62.2 & 68.2 & 51.8 & 86.7\% \\
PDrop~{\scriptsize\textcolor{gray}{(CVPR'25)}}     & 47.5 & 58.8 & 50.5 & 1561 & 55.9 & 69.0 & 69.2 & 50.6 & 82.7\% \\
VisionZip~{\scriptsize\textcolor{gray}{(CVPR'25)}} & 55.1 & 60.1 & 50.4 & 1690 & 77.0 & 69.0 & 72.4 & 55.5 & \textcolor{blue}{92.0}\% \\
\rowcolor{rowblue!40} StepPrune (Ours) & 56.4 & 61.8 & 55.4 & 1698 & 83.1 & 69.3 & 72.9 & 54.7 & \textcolor{red}{94.6\%} \\
\bottomrule
\end{tabular}%
}
\end{table}

StepPrune outperforms every fixed-budget baseline at each operating point, with average recovery of $97.9\%/96.7\%/94.6\%$ at $K{=}192/128/64$. The margin over the strongest competitor, VisionZip, is $+0.1/+0.3/+2.6$ points respectively, and grows as the budget tightens and per-token coverage decisions become harder.

The gain is concentrated on dense and grounded benchmarks such as MME, POPE, and TextVQA, which is consistent with the claim in Sec.~\ref{sec:intro} that conditioning matters most when the prefix carries many mutually dependent cues. POPE is the clearest case: the gap to VisionZip widens from $+0.4$ at $K{=}192$ to $+2.0$ at $K{=}128$ and $+6.1$ at $K{=}64$. This suggests that the independence assumption of one-shot ranking has a measurable impact on grounded behaviour, and that the impact grows as the budget tightens.

\subsection{Results on Qwen2.5-VL-7B}
\label{sec:exp:qwen}

To check whether the conditional policy transfers across backbones, we run the same protocol on Qwen2.5-VL-7B~\cite{qwenvl}, whose vision tower produces a higher native token count and whose language model is trained at a different scale. Table~\ref{tab:main_qwen} reports the results at the same three pruning rates, $66.7\%/77.8\%/88.9\%$.

\begin{table}[t]
  \centering
  \caption{
    Results on Qwen2.5-VL-7B. Block labels, ``Avg.'' definition, and color coding follow Table~\ref{tab:main_llava}.
  }
  \label{tab:main_qwen}
  \scriptsize
  \setlength{\tabcolsep}{4pt}
  \resizebox{0.75\linewidth}{!}{%
  \begin{tabular}{lcccccc}
    \toprule
    Method & MMB & MME & POPE & SQA & VQA$^{\text{Text}}$ & Avg. \\
    \midrule
    \rowcolor{rowgray} \multicolumn{7}{c}{\textit{Upper Bound (100\%)}} \\
    Vanilla & 82.8 & 2304 & 86.1 & 84.7 & 84.8 & 100\% \\
    \midrule
    \rowcolor{rowgray} \multicolumn{7}{c}{\textit{Token Pruning Rate = 66.7\%}} \\
    FastV~{\scriptsize\textcolor{gray}{(ECCV'24)}} & 75.7 & 2072 & 82.2 & 78.5 & 77.9 & 92.3\% \\
    HoloV~{\scriptsize\textcolor{gray}{(NeurIPS'25)}} & 78.3 & 2093 & 85.0 & 79.8 & 78.9 & \textcolor{blue}{94.3}\% \\
    \rowcolor{rowblue!40} StepPrune (Ours) & 80.5 & 2302 & 85.3 & 86.7 & 77.2 & \textcolor{red}{97.9}\% \\
    \midrule
    \rowcolor{rowgray} \multicolumn{7}{c}{\textit{Token Pruning Rate = 77.8\%}} \\
    FastV~{\scriptsize\textcolor{gray}{(ECCV'24)}} & 74.9 & 2036 & 80.7 & 78.0 & 69.0 & 89.2\% \\
    HoloV~{\scriptsize\textcolor{gray}{(NeurIPS'25)}} & 76.5 & 2043 & 82.3 & 79.8 & 70.3 & \textcolor{blue}{90.8}\% \\
    \rowcolor{rowblue!40} StepPrune (Ours) & 78.4 & 2225 & 83.1 & 85.4 & 75.5 & \textcolor{red}{95.5}\% \\
    \midrule
    \rowcolor{rowgray} \multicolumn{7}{c}{\textit{Token Pruning Rate = 88.9\%}} \\
    FastV~{\scriptsize\textcolor{gray}{(ECCV'24)}} & 69.2 & 1940 & 78.6 & 77.4 & 60.3 & 84.3\% \\
    HoloV~{\scriptsize\textcolor{gray}{(NeurIPS'25)}} & 72.4 & 2006 & 80.7 & 79.5 & 61.8 & \textcolor{blue}{87.0}\% \\
    \rowcolor{rowblue!40} StepPrune (Ours) & 75.9 & 2130 & 80.8 & 83.2 & 70.8 & \textcolor{red}{91.9}\% \\
    \bottomrule
  \end{tabular}%
  }
\end{table}

The relative ordering observed on LLaVA-v1.5-7B is preserved on Qwen2.5-VL-7B: at every operating point, StepPrune outperforms both FastV and HoloV~\cite{holov}, with margins of $+3.6$, $+4.7$, and $+4.9$ percentage points over the strongest baseline at the $66.7\%$, $77.8\%$, and $88.9\%$ pruning rates respectively. The advantage therefore extends from the LLaVA-CLIP pair to a backbone with a different vision tower and a different language-model scale, suggesting that the conditional pointer policy is not tied to a specific vision-language pairing.

\subsection{Adaptive Behavior}
\label{sec:exp:adaptive}

\begin{figure}[t]
\centering
\begin{minipage}[t]{0.46\textwidth}
\centering
\includegraphics[width=\linewidth]{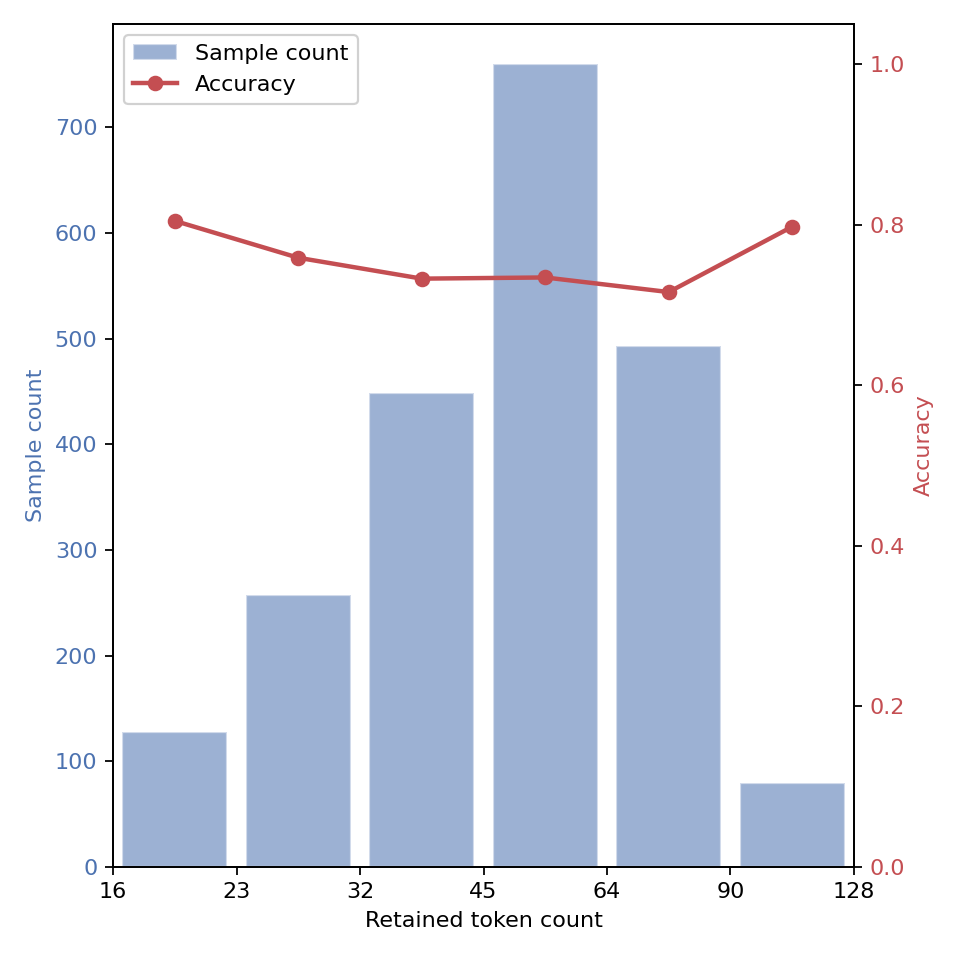}
\caption{Adaptive budget analysis on MME (LLaVA-v1.5-7B). Per-bucket sample count (blue bars, left axis) and average accuracy (red curve, right axis) along the retained token count $K$.}
\label{fig:adaptive_k}
\end{minipage}\hfill
\begin{minipage}[t]{0.50\textwidth}
\centering
\includegraphics[width=\linewidth]{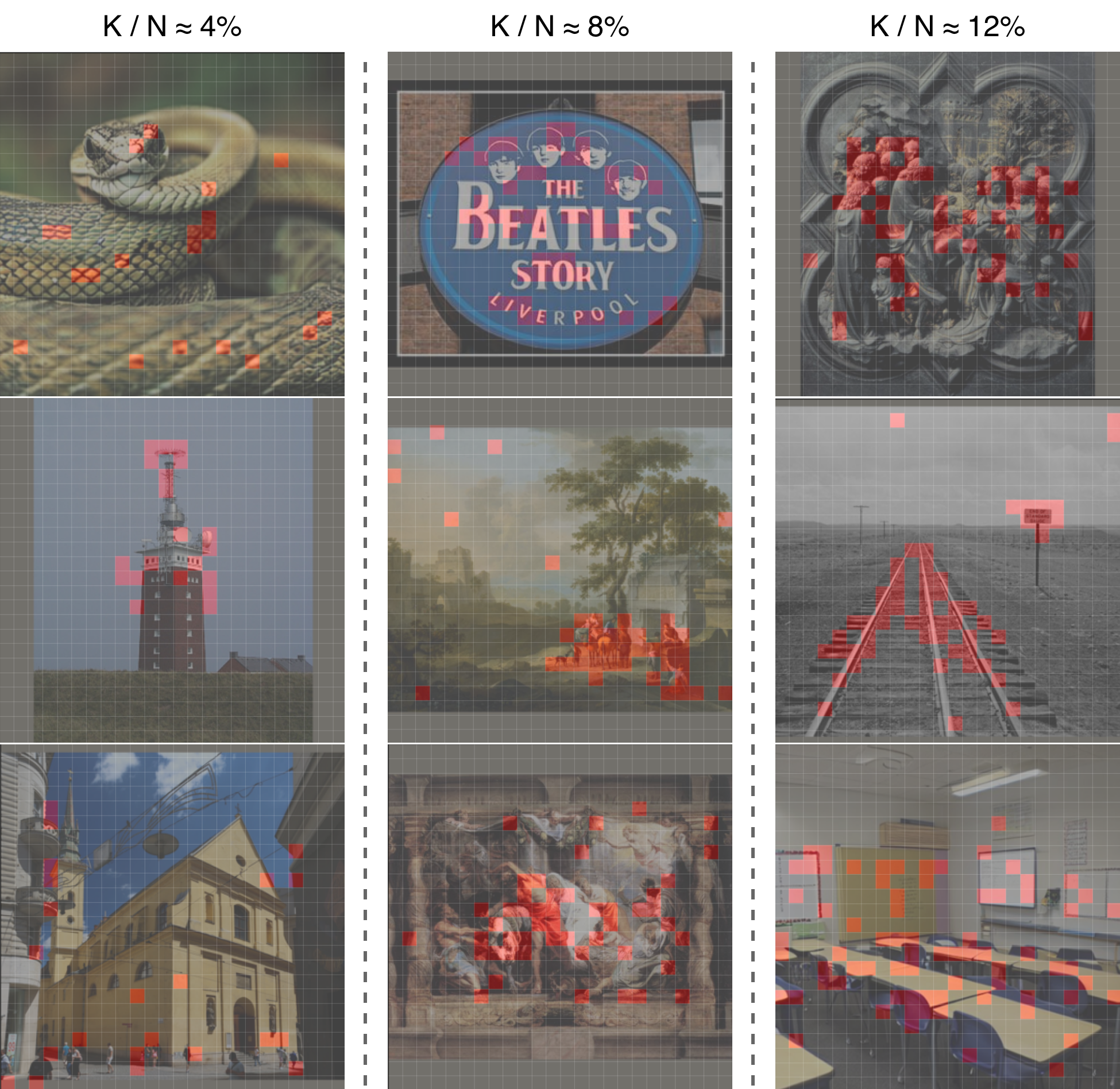}
\caption{Qualitative examples on MME (LLaVA-v1.5-7B). Red overlays mark tokens admitted to $\mathcal{S}$; columns group inputs by retention ratio: $\approx 4\%$ (left), $\approx 8\%$ (middle), $\approx 12\%$ (right).}
\label{fig:qual}
\end{minipage}
\end{figure}

Because $K$ is determined by the model, its distribution across inputs reflects how the policy allocates compute. We examine this on MME with LLaVA-v1.5-7B, binning examples into six log-spaced buckets of $K\in[16,128]$ and plotting per-bucket accuracy on top of the raw histogram (Fig.~\ref{fig:adaptive_k}). A sample of nine inputs grouped by retention ratio is shown in Fig.~\ref{fig:qual}.

The histogram in Fig.~\ref{fig:adaptive_k} is broad: $K$ spans roughly an order of magnitude across MME, with non-trivial mass in every bucket. Despite this spread, the per-bucket accuracy curve is roughly flat. The six bucket means lie within a narrow band, and small-$K$ buckets are not visibly worse than large-$K$ ones. The model does not obtain its small-$K$ wins by sacrificing easier cases, nor does it spend tokens on inputs that a short prefix already serves well; a fixed-ratio pruner cannot reproduce this profile without an external difficulty oracle.

Fig.~\ref{fig:qual} illustrates this adaptivity. Left-column inputs carry a single dominant locus of evidence and the policy halts after a few selections on that locus; right-column inputs lack a dominant subject and the policy keeps selecting until the retained tokens fan out across multiple regions, with the middle column in between. The policy therefore trades selection length for content coverage on a per-image basis.

\subsection{Efficiency Analysis}
\label{sec:exp:efficiency}

A potential concern about the decoder-based design of StepPrune is whether the autoregressive pointer loop offsets the saving from a shorter visual prefix. Table~\ref{tab:efficiency} reports prefill FLOPs and latency on LLaVA-v1.5-7B at $K{=}64$, measured on a single NVIDIA RTX A6000.

\begin{table}[t]
\centering
\caption{Prefill efficiency comparison at $K{=}64$. ``Acc.'' is the average retention from Table~\ref{tab:main_llava}, and the token count is a per-image budget for baselines and the test-split mean $\bar{K}$ for StepPrune.}
\label{tab:efficiency}
\setlength{\tabcolsep}{8pt}
\begin{tabular}{lccccc}
\toprule
Method & Tokens & FLOPs (T) & Latency (ms) & Speed-up & Acc. \\
\midrule
Vanilla (LLaVA)             & 576 & 8.89 & 149.5 & $1.00\times$ & 100.0\% \\
FastV~\cite{fastv}          & 64  & 2.39 & 102.6 & $1.46\times$ & 76.8\% \\
PDrop~\cite{pdrop}          & 64  & 4.89 & 105.9 & $1.41\times$ & 82.7\% \\
\rowcolor{rowblue!40} StepPrune (Ours) & 64  & 2.28 & 79.5  & $\mathbf{1.88\times}$ & \textbf{94.6\%} \\
\bottomrule
\end{tabular}
\end{table}

StepPrune attains the lowest FLOPs and prefill latency among the compared methods, achieving a $1.88\times$ speed-up at $94.6\%$ accuracy retention. Although the pointer decoder introduces additional parameters, the modules trained in Sec.~\ref{sec:method} remain small relative to the 7B language model and are invoked only once before the LLM forward pass. Per-layer LLM cost grows superlinearly with the prefix length, so trimming from $576$ to $64$ tokens saves far more compute than the policy itself spends. The progressive in-layer dropping of PDrop, by contrast, repeatedly enters the language model and inflates FLOPs accordingly, capping its wall-clock gain.

\subsection{Ablation Studies}
\label{sec:exp:ablation}

We ablate two design choices of StepPrune on LLaVA-v1.5-7B and, following the protocol of AutoSelect~\cite{autoprune}, report the average performance retention over four benchmarks: GQA, MMB, POPE, and MME. Table~\ref{tab:ablation} reports both studies in a single table; each block varies one design dimension while holding the other settings at their defaults (marked with $^{\dagger}$).

\begin{table}[t]
\centering
\caption{Ablation studies on LLaVA-v1.5-7B. Each row reports the average performance retention (\%) over GQA, MMB, POPE, and MME. Rows marked with $^{\dagger}$ indicate the default configuration used elsewhere in the paper.}
\label{tab:ablation}
\setlength{\tabcolsep}{8pt}
\begin{tabular}{lc}
\toprule
Variant & Avg.\ Acc. \\
\midrule
\rowcolor{rowgray}\multicolumn{2}{l}{\textit{(a) Encoder self-attention direction}} \\
\quad Causal (unidirectional)                                  & 92.4 \\
\quad Bidirectional$^{\dagger}$                                & 94.6 \\
\midrule
\rowcolor{rowgray}\multicolumn{2}{l}{\textit{(b) Scorer aggregation (see Eq.~\ref{eq:soft_score})}} \\
\quad Last-step pointer only                                   & 90.3 \\
\quad Uniform sum ($A_{t}{\equiv}1,\,\beta{=}1$)               & 83.9 \\
\quad Alive-weighted only ($\beta{=}1$)                        & 89.5 \\
\quad $A_{t}\cdot p_{t}(i)\cdot\beta^{t}$$^{\dagger}$          & 94.7 \\
\bottomrule
\end{tabular}
\end{table}

\paragraph{(a) Encoder self-attention direction.}
The input encoder $\mathrm{Enc}$ is not autoregressive: every image is observed in full before selection begins, so there is no principled reason to mask it causally, although a causal default is common in transformer code. Replacing the bidirectional encoder with a causal one removes information from later patch positions in the scan order and costs $2.2$ points of average accuracy. This confirms that bidirectional context over the whole image is needed for selection quality.

\paragraph{(b) Scorer aggregation.}
The soft score $s_i$ that feeds the noise-gating surrogate combines three terms (Eq.~\ref{eq:soft_score}): the alive probability $A_{t}$, the pointer probability $p_{t}(i)$, and a geometric depth decay $\beta^{t}$. Removing any one of them, by using only the last-step pointer, dropping the alive weight, or setting $\beta{=}1$, costs $4.4$--$10.8$ points of accuracy, because the degenerate aggregation fails to translate a full selection trajectory into a consistent per-token credit signal.

\section{Conclusion}
\label{sec:conclusion}

StepPrune replaces score-and-top-$K$ with a sequential alternative: a pointer-style decoder selects one token per step conditioned on prior choices, and a learned stop action $\varnothing$ produces the retained subset $\mathcal{S}$ and its size $K$ jointly. A variance-preserving noise gate carries gradients through the discrete argmax for end-to-end training under the language-modelling loss. On LLaVA-v1.5-7B and Qwen2.5-VL-7B, StepPrune preserves $97.9\%/96.7\%/94.6\%$ of the full-model accuracy at $66.7\%/77.8\%/88.9\%$ pruning rates and yields a $1.88\times$ prefill speed-up at the most aggressive setting, and the distribution of $K$ tracks input complexity rather than collapsing onto a fixed budget.



\bibliographystyle{waica}

\end{document}